\documentclass{bmvc2k}

\title{Gradual Channel Pruning while Training using Feature Relevance Scores for Convolutional Neural Networks}

\addauthor{Sai Aparna Aketi}{saketi@purdue.edu}{1}
\addauthor{Sourjya Roy}{roy48@purdue.edu}{1}
\addauthor{Anand Raghunathan}{raghunathan@purdue.edu}{1}
\addauthor{Kaushik Roy}{kaushik@purdue.edu}{1}

\addinstitution{
 Electrical and Computer Engineering\\
 Purdue University\\
 West Lafayette, Indiana, USA.
}

\runninghead{S.A.Aketi, S.Roy, A.Raghunathan, K.Roy}{Gradual Channel Pruning}

\begin{document}

\maketitle

\begin{abstract}
The enormous inference cost of deep neural networks can be scaled down by network compression. Pruning connections is one of the predominant approaches used for deep network compression. However, existing pruning techniques have one or more of the following limitations: 1) Additional energy cost on top of the compute-heavy training stage due to pruning and fine-tuning stages, 2) Layer-wise pruning based on the particular layer’s statistics, ignoring the effect of error propagation in the network, 3) Lack of efficient estimate for determining the global importance of channels,
4) Unstructured pruning may not lead to any energy advantage when implemented in a GPU and may require specialized hardware to reap the benefits. To address the above issues, we present a simple-yet-effective \textit{gradual channel pruning while training} methodology using a novel data-driven metric referred to as \textit{feature relevance score}. The proposed technique gets rid of the additional retraining cycles by pruning the least important channels in a structured fashion at fixed intervals during the regular training phase. Feature relevance scores help in efficiently evaluating the contribution of each channel towards the discriminative power of the network. We demonstrate the effectiveness of the proposed methodology on architectures such as VGG and ResNet using datasets such as CIFAR-10, CIFAR-100 and ImageNet, and successfully achieve significant model compression while trading off less than $1\%$ accuracy. 
\end{abstract}

\section{Introduction}
\label{sec:intro}
Convolutional Neural Networks have achieved great success in a variety of computer vision tasks. The advantages comes from the deeper architectures with millions of parameters \cite{NIPS2012_4824}. Such large models require immense computational power and memory for storage in both the training and the testing phases. As a result, deep learning models are unsuitable for applications with limited resources, such as the edge devices. However, recent studies suggest that the deep neural networks have significant redundancy \cite{prakash2018repr, redundancy} and network pruning can be used for model compression, leading to much reduced computational cost and memory usage without significant degradation in performance. 

Most of the existing pruning techniques usually involve the following three stages \cite{rethinking_iclr19}: 1) Training an over-parameterized network till convergence, 2) Pruning based on a predefined criteria, 3) Fine-tuning/ re-training to regain the accuracy. The main limitation with such three stage process of existing pruning methods is that the pruning and fine-tuning stages, iterative in most cases \cite{dcp_nips18, iccv17_slimming, thinnet, connections_nips15, cvpr_2019}, impose additional computation (hence, time and energy) requirements on top of the compute-heavy training stage. 
Based on the structure and criteria used for pruning, most of the previous works on network pruning also suffer from one or more of the following problems: 
(a) Unstructured pruning methods result in sparse weight matrices which are unstructured and require dedicated hardware/libraries for compression and speedup \cite{Eie, rethinking_iclr19}. Examples of unstructured pruning methods include \cite{guo_nips16, lebedev_cvpr16, dropout_icml17, lee2018snip, christos_iclr18}.
(b) Predefined structured pruning, where the pruned target architecture is defined by the user, does not have any advantage over training the targeted model from scratch \citep{rethinking_iclr19}. Examples of predefined structured pruning methods include \cite{soft_2018, nisp, pruning_iclr17, iccv17_slimming, channel_iccv}.
(c) Pruning neurons layer-by-layer either independently \cite{compressing_iclr16} or greedily \cite{pruning_iclr17, thinnet, dcp_nips18}, without jointly considering the statistics of all layers, can lead to significant reconstruction error propagation \cite{nisp}.
(d) Use of an iterative metric (which requires an optimization step in a few cases) to evaluate the importance of channels or nodes is computationally expensive \cite{dcp_nips18}.

In this paper, we aim to overcome all the above mentioned drawbacks. First, we adopt the idea that it is not necessary to train the model till convergence before pruning \cite{yue2019really, roy2020}. Based on this idea, we merge the training and pruning stages, eliminating the fine-tuning stage which reduces the additional computational requirements. We prune a fixed, predefined percentage of least important channels from the entire network after every few epochs during the training phase. It eliminates the disadvantages of unstructured and predefined structured pruning. Another key factor in pruning is evaluating the importance of channels. We compute the importance of each channel in the network using a data-driven metric that we call \textit{feature relevance score}. The contribution of each channel towards the discriminative power of the neural network can be evaluated efficiently through feature relevance scores. Feature relevance scores consider the statistics of the entire model, reducing the reconstruction error propagation. 

Feature relevance score is computed based on normalized class-wise accuracy of the model and class-wise relevance score of each channel. The class-wise relevance scores are computed using a technique called \textit{Layer-wise Relevance Propagation (LRP)} proposed in \cite{lrp:2015}. 
The relevance score of a channel for any given class in the dataset indicates its average contribution in activating the output node corresponding to that class, and is given by the weighted average of the relevance scores of all classes in the dataset for a given channel where the weights are determined by the class-wise accuracy of the model. The score of a channel denotes how important the channel is in making the final prediction and is not based on statistics of individual layers. 
Feature relevance scores are computed recursively using the training data, where the true labels are used to determine the scores at the final output layer and one backward pass is made through the network to determine the relevance scores of the channels in the remaining layers.
Thus, determining feature relevance scores is not computationally expensive and the computational complexity is same as 1-2 training epochs as shown in section.~\ref{effort}.

We evaluate the efficiency of our methodology on CIFAR-10, CIFAR-100 \cite{cifar:2009} and ImageNet \cite{imagenet} using standard CNN architectures such as VGGNet \cite{vgg16} and ResNet \cite{res:2016}. Our experiments show that the proposed technique can prune 59\% of the parameters resulting in 56\% reduction of FLOPS (Floating Point Operations) for ResNet-110 on CIFAR-10 with an accuracy drop of 0.01\%. On ImageNet, when pruning 25\% of channels of ResNet-34, we observe an accuracy drop of 1.5\%. This shows that the proposed technique is able to achieve significant compression (comparable to the existing pruning techniques) while reducing the additional energy cost required by the existing pruning methods.

\subsection{Contribution}
We introduce a novel data-driven metric called \textit{feature relevance score} which computes the importance of each channel in the entire network by efficiently propagating the importance scores from the final output layer. This metric is utilized in \textit{gradual channel pruning while training} where we gradually prune channels from the entire network (automatic structured pruning) at fixed intervals over the regular training phase, reducing computational and time complexity.

\section{Related work}
Deep model compression has been successfully achieved through network pruning techniques which remove the redundant / unnecessary connections. Network pruning can be done at various granularity levels such as individual weights, nodes, channels or even layers. The pruning techniques which prune at the level of individual weights or nodes can be categorized under unstructured pruning methods. Some of the recent works such as \cite{connections_nips15, guo_nips16, lebedev_cvpr16, christos_iclr18} have proposed unstructured pruning. The authors of \cite{connections_nips15} proposed iterative weight pruning where the weights with magnitude less than a given threshold are removed. 
The authors in \cite{lebedev_cvpr16, christos_iclr18} explore sparsity regularizers to accelerate the deep neural networks.
However, the disadvantage of unstructured pruning techniques is that they do not lead to realistic speed ups and compression without dedicated hardware \cite{Eie}.

The pruning techniques which prune at the level of channels or layers come under structured pruning methods. 
The important step in channel pruning is evaluating the importance of channels. Network trimming technique proposed in \cite{hu_network} prunes the channels based on average percentage of zeros (APoZ). The importance of channels is determined by computing the sum of absolute value of weights in each channel in \cite{pruning_iclr17}. Some recent works such as \cite{iccv17_slimming, Ye_iclr18}, include channel-wise scaling factors with sparsity constraints during training, whose magnitudes are then used for channel pruning.
Reconstruction methods which transform the channel selection problem into an optimization of reconstruction error with consideration of efficiency have been proposed in \cite{thinnet, channel_iccv, dcp_nips18}. However, all the above techniques require additional pruning and re-training effort.

The authors in \cite{Garg_2020, lee2018snip} have proposed one-shot pruning which aims to avoid expensive prune-retrain cycles. The technique proposed in \cite{lee2018snip} prunes the connections after initialization in a data-dependent way which makes it sensitive to the weight-initialization. This technique falls under unstructured pruning as it prunes individual weights. The authors in \cite{Garg_2020} prune the pre-trained network in one-shot. But the pruned network has to be trained from scratch which still requires a considerable amount of computation.  
Note, pruning while training has been proposed in \cite{yue2019really, roy2020}. These techniques prune the channels based on the sum of their absolute weights ($L_1$--norm) or sum of the square of their weights ($L_2$--norm).
However, $L_1$ or $L_2$--norms do not consider the effect of the channel on the final output as they only depend on the weights of a particular channel.
In this paper, we explore gradual channel pruning while training using a novel data-driven metric \textit{feature relevance scores}. 
Feature relevance scores are computed by back-propagating the output of winner output node and redistributing it to previous layers using a pre-defined rule based on activations and strength of the connectivity. Hence, feature relevance scores consider the statistics of the entire network and efficiently evaluate the discriminative power of each channel in the network.

\section{Methodology}
In this section, we present our approach to evaluate the importance of channels and how we gradually prune them during training. 

\subsection{Feature Relevance Scores}
\label{sec:frs}
CNNs trained for classification tasks compute a set of features at each layer. The contribution of each feature-map (or channel) in the network to a given prediction depends on its activation values and their propagation to the final layer. 
\begin{algorithm}[ht]
\textbf{Input:} CNN, Training data $\{(x_i,y_i)\}_{i=1}^N$, class-wise accuracy: $acc$\\
\textbf{Parameters:} number of classes: $c$,\hspace{1mm} number of layers: $L$,
feature maps at layer $l$: $\{f_1,\dots, f_r\}$,\hspace{1mm} relevance score of node $p$ at
layer $l$ = $R_p^l$, feature relevance scores of layer $l$: $FS_l$ \\
1. Initialize feature-relevance matrix for given layer \textit{l}: $FM_l=zeros(c,r)$\\
2. \textbf{for} each sample $(x_i,y_i)$ in training data \textbf{do}\\
3.\hspace{0.5cm}Forward propagate the input $x_i$ to obtain the activations of all nodes in the DNN \\
4.\hspace{0.5cm}Compute relevance scores for output layer: $R_p^L = \delta(p-y_i)\hspace{2mm} \forall p\in \{1,\dots,c\}$\\
.\hspace{6cm}$\delta(p-y_i) =$ Kronecker delta function \\
5.\hspace{0.5cm}\textbf{for} $k$ in $range(L-1,l,-1)$ \textbf{do}\\
6.\hspace{1cm}Back propagate relevance scores:  $R_p^k = \sum_{q}(\alpha \frac{(a_p w_{pq})^{+}}{\sum_p {(a_p w_{pq})^{+}}} -\beta\frac{(a_p w_{pq})^{-}}{\sum_p {(a_p w_{pq})^{-}}})R_q^{k+1}$\\ .\hspace{1.2cm}$\forall$ $p \in$ nodes of layer $k$, $\alpha - \beta = 1$,  $a_p =$ activations,  $w_{pq}=$ weights, \\
.\hspace{1.2cm}$(a_p w_{pq})^+$: positive weight components, $(a_p w_{pq})^-$: negative weight components.\\
7.\hspace{0.5cm} \textbf{end for}\\
8.\hspace{0.5cm} Compute average relevance score per feature map \\
9.\hspace{0.5cm} Relevance score vector at layer $l$: $R^l = \big\{R^l_{f_j} =\frac{1}{\sum\limits_{p \in f_j}1} \big(\sum\limits_{p \in f_j} R_p^l\big)\big\}_{j=1}^{r}$\\
10.\hspace{0.5cm}Update feature-relevance matrix: $FM_l(y_i,:) += R^l $\\
11. \textbf{end for}\\
12. Normalize rows of feature-relevance matrix: $FM_l(p,:) = \frac{1}{\sum\limits_{\forall y_i\in p}1}* FM_l(p,:)$ $\forall p \in [1,c]$\\
13. Compute normalized class-wise accuracy: $\lambda_p = acc_p/max(acc), \hspace{1mm}v_p=\frac{1}{\lambda_p}$ $\forall p \in [1,c]$\\
14. Compute feature relevance scores: $FS_l = \frac{1}{v}\sum\limits_{p=1}^{c}v_p*FM_l(p,:)$ \hspace{1mm} where $ v = \sum\limits_{p=1}^{c}v_p$\\
15. \textbf{return} feature relevance scores $FS_l$ of layer $l$
\caption{Methodology to Compute Feature Relevance Scores $FS_l$ of Layer $l$}
\end{algorithm}
For a given instance, \textit{Layer-wise Relevance Propagation (LRP)} proposed in \cite{lrp:2015} determines the contribution of each node towards the final prediction. LRP allocates a relevance score to each node in the network using the activations and weights of the network for an input image.
The relevance scores at output nodes are determined based on true label of an instance. For any input sample $(x_i,y_i)$, the output node corresponding to true class, $y_i$, is given a relevance score of $1$ and the remaining nodes get a score of $0$. 
The relevance scores of output nodes are then back propagated based on $\alpha \beta$-\textit{decomposition rule} \cite{lrp:2016} with $\alpha=2$ and $\beta=1$. More details in $\alpha \beta$-\textit{decomposition rule} is provided in the supplementary material. 

Algorithm 1 shows the pseudo code for computing the feature-relevance score of layer $l$.
After determining the relevance scores at each node in the network using LRP, we compute the relevance score of every feature map $f_i$ at layer $l$ by averaging the scores of all nodes corresponding to $f_i$. The relevance vector of a feature map $f_i$ is obtained by taking class-wise average over relevance scores of all training samples and forms the $i^{th}$ column of feature-relevance matrix ($FM_l$).
The weighted average of rows of $FM_l$ with class-wise accuracy returns feature relevance scores ($FS_l$) for any layers $l$.
The computed feature relevance scores are then utilized to determine the least important channels that can be pruned.

\subsection{Gradual Channel Pruning while Training}

The proposed gradual channel pruning while training technique is shown in Algorithm 2. It uses feature relevance scores (ref.~\ref{sec:frs}) to evaluate the importance of each channel.
The pruning methodology merges the pre-training and the pruning phases reducing the additional efforts required by the traditional pruning techniques. Starting from scratch, pruning takes place after every few epochs of the actual training phase. At each pruning stage, the least important $x$ channels are removed.  
\begin{algorithm}[ht]
\textbf{Input:} CNN, Training data $\{(x_i,y_i)\}_{i=1}^N$\\
\textbf{Parameters:} prune $x$ channels after every $n$ epochs till $N1$ epochs, maximum epochs: $N$. \\
1. Initialize the CNN model\\
2. \textbf{for} epoch = $1,2,\dots,N$\\
3.\hspace{0.5cm}Update the learning rate if required \\
4.\hspace{0.5cm}Train the CNN using SGD algorithm with momentum and weight decay for an epoch\\
5.\hspace{0.5cm}\textbf{if} epoch$\%n==0$ and epoch$<N1$:\\
6.\hspace{1.2cm}Determine class-wise accuracy of training data.\\
7.\hspace{1.2cm}Compute feature relevance scores of CNN with current weights using algorithm 1 \\
8.\hspace{1.2cm}Prune $x$ channels which have least feature relevance scores globally.\\
9.\hspace{0.5cm}\textbf{end if}\\
10. \textbf{end for}\\
11. \textbf{return} Pruned CNN model
\caption{Gradual Pruning while Training}
\end{algorithm}
We do not prune during the last few epochs of training allowing the model to converge. In particular, say the number of training epochs for the model is $N$ and the pruning interval is defined as $(1,N1)$ epochs. The model is pruned after every few epochs (say $n$) of training during the pruning interval i.e., after epoch $n, 2n,\dots, kn$ where $k=int(\frac{N1}{n})$. The model is not pruned in the training interval $(N1,N)$.

The proposed pruning technique introduces the following hyper-parameters: The number of channels to be removed at each pruning stage ($x$), the number of training epochs between two consecutive pruning stages ($n$) and $N1$ indicating the end of pruning interval. Note $x$ is user defined and determines the required final pruning percentage.
The total number of channels pruned is equal to $k*x$ where $k$ is the number of times the model is pruned and is given by $int(\frac{N1}{n})$.
Here, $N$ is same as the number of epochs required to train the baseline CNN model till convergence. 

\section{Experimental Results}
In this section, we evaluate the effectiveness of the proposed pruning methodology. The time and computational complexity of the proposed pruning technique is also presented.
We compare our technique with several existing and state-of-the-art pruning methods such as \cite{He_2019_CVPR, nisp, soft_2018, cvpr_2019, NIPS2019_gd, dcp_nips18}. We have used three different datasets namely CIFAR-10, CIFAR-100 and ImageNet and trained them on VGGNet and ResNet architectures. For all the datasets and architectures, we report the percentage drop in accuracy, the percentage drop in parameters and the percentage reduction in FLOPS (Floating Point Operations) during inference with respect to the baseline unpruned model. We also report if pre-trained model is required as initialization and the number of additional epochs required for fine-tuning the pre-trained model after pruning. The implementation details are provided in the supplementary material. 

\subsection{Pruning Effort}
\label{effort}
In this section, we discuss the time and computational complexity of computing feature-relevance scores which determine the channel importance.
The time complexity of the feature relevance score computation is reported in the form of \textbf{search time}. 
Search time represents the time required to determine the $x$ channels with least feature relevance scores that have to be pruned at a pruning stage (step 6-8 in Algorithm 2). The value of $x$ used to compute the search time in Table.~\ref{tab:effort} corresponds to the maximum pruning percentage reported in Table.~\ref{tab:vgg_res}.
The computational complexity is reported in terms of \textbf{effort factor ($\rho$)}. Effort factor is defined as the ratio of number of FLOPs required to compute feature relevance scores to the number of FLOPs required for one training epoch (see Equation.~\ref{eq:1}). The number of FLOPs required for a training epoch is considered to be three times the number of FLOPs required for a forward-pass \cite{ankit2019panther}.
The experiments were conducted on a system with a Nvidia GTX 1080ti GPU.

\begin{equation}
\label{eq:1}
    \rho = \frac{\# \hspace{2mm} of \hspace{2mm} FLOPs(feature \hspace{2mm} relevance \hspace{2mm} scores)}{\#\hspace{2mm}  of \hspace{2mm}  FLOPs(forward+backward \hspace{2mm} pass)}
\end{equation}

We have reported the the search time for the first pruning step which is the upper bound for the remaining pruning steps as the complexity of the model reduces with each pruning step (refer Figure.~\ref{fig:vgg_params}(b)). The additional time required to obtain the pruned model as compared to training an unpruned model is bounded by the search time (see Table.~\ref{tab:effort}) multiplied by the number of pruning steps.
Also, for large datasets such as ImageNet, we have used a random subset of training data (around 0.08 million images) to compute the feature relevance scores rather than the entire dataset which has a million images. 
The computation of feature relevance scores requires the access to activations at each layer. We have used the inbuilt function, $register\_forward\_hook$, from PyTorch to obtain the hidden layer activations which is not compatible with data parallelism. Hence, the reported search time does not include the benefits of data parallelization across multiple devices within a GPU.

\begin{table}[H]
\caption{Effort required by the proposed pruning methodology. Both the search time and effort factor are computed for the first pruning step.}
\label{tab:effort}
\vskip 0.15in
\begin{center}
\begin{small}
\begin{tabular}{|l|ccc|}
\hline
Dataset& Model & search time (mins) & Effort factor ($\rho$)\\
\hline
  & VGG-16& 2.24
  & 1.29\\
  CIFAR-10& ResNet-56& 2.67 & 1.31 \\
  &ResNet-110& 4.61 & 1.32\\
  &ResNet-164& 7.76 & 1.67\\
 \hline
ImageNet& ResNet-34 &20.53 &1.36 \\
  \hline
 \end{tabular}
\end{small}
\end{center}
\vspace{-0.5cm}
\end{table}

The effort factor of the proposed pruning technique ranges from $(1.3-1.7)$ as shown in Table.~\ref{tab:effort}. This shows that the computational complexity of computing feature relevance scores (i.e., a pruning step) is less than the computational complexity of two training epochs. Note that we have reported the effort factor for the first pruning step which is the upper bound for the remaining pruning steps (refer Figure.~\ref{fig:vgg_params}(b)). In our experiments, the number of pruning steps for the CIFAR-10 dataset is set as $7$ and this results in the upper bound of the total pruning effort to be equivalent to $9-12$ training epochs. 
Also, the computational complexity required per training epoch decreases with each pruning step as compared to the training of unpruned model.
Hence, the total computational complexity of training the model from scratch and pruning it through the proposed methodology is less than that of training an unpruned model till convergence.

\subsection{Comparisons on CIFAR-10 and CIFAR-100}
\begin{figure}[ht]
\begin{tabular}{ccc}
\bmvaHangBox{\fbox{\includegraphics[width=5.6cm]{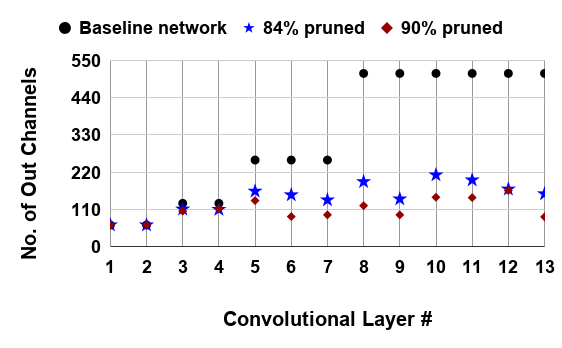}}}&
\bmvaHangBox{\fbox{\includegraphics[width=5.6cm]{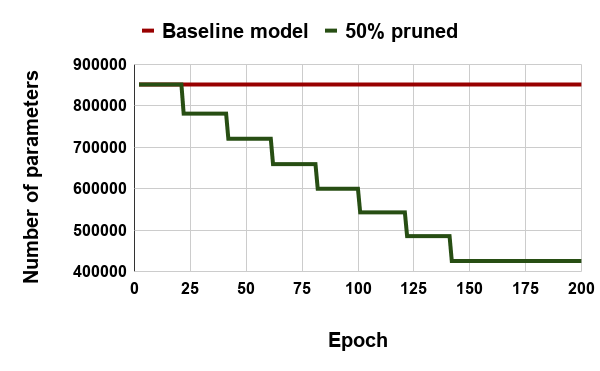}}}\\
(a)&(b)
\end{tabular}
\caption{(a) The number of channels pruned in each convolutional layer for VGG-16 network trained on CIFAR-10 dataset. (b) Variation in number of parameters during the training phase of the pruning methodology for ResNet-56 architecture trained on CIFAR-10 dataset.}
\label{fig:vgg_params}
\end{figure}

Tables.~\ref{tab:vgg_res} show the comparison of the proposed technique with various existing pruning techniques for CIFAR-10 and CIFAR-100 datasets.
The number of channels pruned at each layer for CIFAR-10 dataset trained on VGG-16 is shown in figure.~\ref{fig:vgg_params}(a). We observed that the percentage of pruning is higher in the latter layers than the initial layers. For the CIFAR-10 dataset, the experimental results show that our technique was able to prune $(55-90)\%$ of the chosen baseline network with an accuracy drop of less than $1\%$ as compared to the unpruned model (refer table.~\ref{tab:vgg_res}).
We were able to achieve $59\%$ pruning resulting in $56\%$ reduction in FLOPs for ResNet-110 architecture trained on the CIFAR-10 dataset with an accuracy drop of $0.01\%$ as compared to the unpruned model. The pruning percentage achieved for ResNet architectures trained on the CIFAR-100 dataset varied from $(20-30)\%$ resulting in $(30-40)\%$ reduction in FLOPs (refer table.~\ref{tab:vgg_res}). 
The pruning percentage of the CIFAR-100 is less because of the complexity of the dataset. Note that for ResNet-110 on CIFAR-10, the class relevance scores (refer sec.~\ref{sec:frs}) were equally weighted instead of using class-wise accuracies. We have observed that weighted class relevance scores according to class-wise accuracies resulted in smaller accuracy drop in case of more complicated datasets such as CIFAR-100.

\begin{figure}[ht]
\begin{tabular}{ccc}
\bmvaHangBox{\fbox{\includegraphics[width=5.6cm]{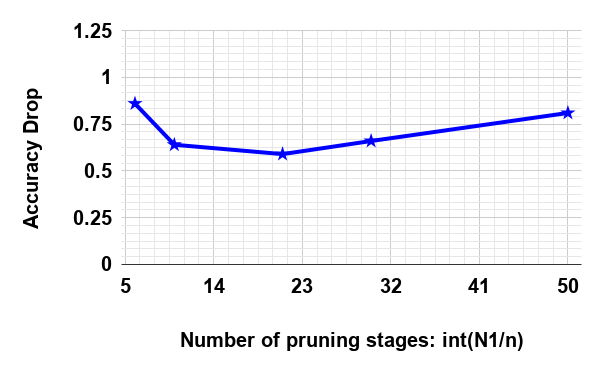}}}&
\bmvaHangBox{\fbox{\includegraphics[width=5.6cm]{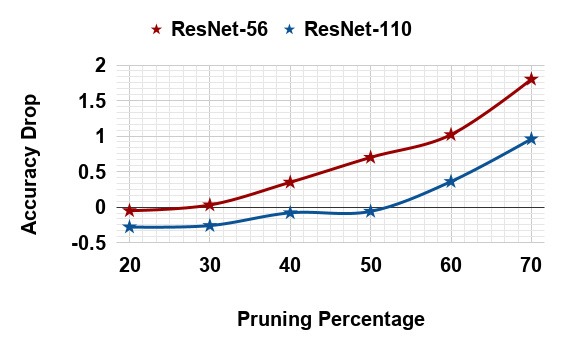}}}\\
(a)&(b)
\end{tabular}
\caption{(a) Accuracy drop vs number of pruning stages for ResNet-56 architecture trained on CIFAR-10 dataset. (b) Accuracy drop vs pruning percentage for ResNet architecture trained on CIFAR-10 dataset.}
\label{fig:n}
\end{figure}

In the proposed methodology, the total number of training epochs remain same as that of the baseline model and the number of trainable parameters decrease periodically as shown in Figure.~\ref{fig:vgg_params}(b). Hence, the computational complexity required per training epoch decreases compared to the baseline model as the number of epochs increases.  
The accuracy drop increases with increase in the number of pruning stages $k$ as shown in Figure.~\ref{fig:n}(a).
Figure.~\ref{fig:n}(b) indicates the percentage of accuracy drop as we increase the pruning percentage by changing the hyper-parameter $x$. 
We observe that the accuracy drop increases as the pruning percentage increases which is as expected. 

\begin{table}[ht]
\caption{Comparison of pruning VGG-16 and ResNet architectures trained on CIFAR-10 and CIFAR-100.}
\label{tab:vgg_res}
\vskip 0.15in
\begin{center}
\begin{small}
\begin{tabular}{|l|ccccccc|}
\hline
Dataset & Network & Method & \% Acc. & \% Params & \% FLOPs& Pre-& Add.\\
& & & Drop & Drop & Reduction& trained? & epochs\\
\hline
  & & \cite{dcp_nips18}& \textbf{-0.58} & \textbf{93.5} & 65.0& Y & 400\\
  CIFAR10& VGG-16& \cite{Garg_2020} & 0.71 & 87.0 &65.5 & Y & 164\\
  && \cite{roy2020} &1.30 & 80.0& - & N & 0\\
  & & ours &1.00 & 90.5& \textbf{65.6}& N &0\\
\hline
  & & \cite{NIPS2019_gd}&\textbf{-0.33} &53.3  & \textbf{60.1}& Y &140\\
   CIFAR-10 & ResNet-56 & \cite{dcp_nips18}& -0.01 & \textbf{70.4} & 47.1& Y & 400\\
   & & \cite{He_2019_CVPR}&0.10 & 40.0&52.6&Y&200 \\
  & & ours & 0.72 & 52.8 &52.1& N & 0\\
 \hline
  & & \cite{pruning_iclr17}& 0.20& 32.4 &38.6&Y&40\\
  & & \cite{dong_cvpr17}& 0.19&-&34.2&N&0\\
  CIFAR-10 &ResNet-110 & \cite{nisp}&0.18 & 43.3 & 43.8& Y&-\\
 & & \cite{He_2019_CVPR}& \textbf{-0.16}& 40.0&52.3& Y&200\\
  & & ours & 0.01&\textbf{58.8} & \textbf{55.7}& N& 0\\
 \hline
  CIFAR-10& ResNet-164 & \cite{iccv17_slimming}&\textbf{-0.15} & 35.2& 44.9& Y & 160\\
  & & ours & 0.54 & \textbf{60.2} & \textbf{70.3}& N & 0\\
  \hline
  CIFAR-100& ResNet-56 & ours & 0.67&30.2&32.9& N & 0\\
 \hline
 CIFAR-100& ResNet-110 & ours& 0.93&22.6&39.0& N & 0\\
 \hline
 & & \cite{iccv17_slimming}&0.54 & \textbf{29.7}& \textbf{50.6}&Y& 160\\
 CIFAR-100 & ResNet-164 &ours& \textbf{0.44} & 26.2 &39.1& N & 0\\
\hline
\end{tabular}
\end{small}
\end{center}
\vspace{-0.5cm}
\end{table}

\subsection{Comparisons on ImageNet}
To demonstrate the effectiveness of the proposed method on large-scale datasets, we further apply our method on ResNet-34 on ImageNet dataset which has 22 million parameters. The baseline accuracy of ResNet-34 on ImageNet dataset is $75.12\%$. We were able to prune $25\%$ of the network with an accuracy drop of $1.5\%$ as compared to the unpruned model. Note that we do not prune the parameters in the linear layer of ResNet-34.  
\begin{table}[H]
\caption{Comparison of pruning ResNet-34 on ImageNet.}
\label{tab:imagenet}
\vskip 0.15in
\begin{center}
\begin{small}
\begin{tabular}{|l|cccccc|}
\hline
Network& Method & \% Acc. & \% Params & \% FLOPs & Pre-trained? & Additional\\
& & Drop & Drop & Reduction &Y/N & epochs\\
 \hline
 & \cite{pruning_iclr17} & 1.11 & 10.8 & 24.2& Y&20\\
  ResNet-34& \cite{He_2019_CVPR} & 1.29 & 30.0&41.1&Y&60\\
  & \cite{cvpr_2019} & 0.48 & 21.1 & 22.2& Y & 25\\
  & ours & 1.48& 24.7 & 15.9&N&0\\
  \hline
 \end{tabular}
\end{small}
\end{center}
\vspace{-0.5cm}
\end{table}

\section{Discussion}

Unlike existing pruning techniques, the proposed technique does not require the model to be trained till convergence before pruning. 
Instead, the model is pruned globally in a structured fashion during the actual training phase. This reduces the computational and time complexity of training along with inference. 
Hence, the proposed gradual channel pruning technique will allow the training on edge devices (with limited resources) which might not be feasible with the existing pruning techniques. 
The pruning methodology utilizes a data driven metric, \textit{feature relevance scores}, to determine the redundant or less important channels. 
Replacing feature relevance score with much simpler metrics such as $L_1$ or $L_2$ norm can lead to better computational complexity. However, these metrics only consider the statistics of individual layers, ignoring the effect of error-propagation in the network and result in higher accuracy drop after pruning. 
Feature relevance scores are computed using Layer-wise Relevance Propagation which is state-of-the-art explainable technique.
It utilizes the training data to determine the average effect of the channels on the predicted output. In particular, feature relevance scores include the statistics of the entire network by considering the activations of each channel and its propagation path to the final layer. Hence, the proposed technique uses a better metric to quantify the contribution of a channel towards the discriminative power of the network without an optimization step. 

\section{Conclusion}
Convolutional Neural Networks are crucial for many computer vision tasks and require energy efficient implementation for low-resource settings. In this paper, we present a gradual channel pruning technique while training for CNN compression and acceleration based on feature relevance scores. The channel importance is efficiently evaluated using feature relevance scores after every few epochs during training and the least important channels are pruned. 
The proposed pruning methodology is free of iterative retraining, which reduces the computational and time complexity of pruning a deep neural network.  
The effectiveness of the our pruning technique has been demonstrated using benchmark datasets and architectures. We observe that the proposed technique is able to achieve significant compression and acceleration with less than $1\%$ loss in accuracy. 

\bibliography{egbib}

\begin{thebibliography}{38}
\providecommand{\natexlab}[1]{#1}
\providecommand{\url}[1]{\texttt{#1}}
\expandafter\ifx\csname urlstyle\endcsname\relax
  \providecommand{\doi}[1]{doi: #1}\else
  \providecommand{\doi}{doi: \begingroup \urlstyle{rm}\Url}\fi

\bibitem[Alex and Geoffrey(2009)]{cifar:2009}
Krizhevsky Alex and Hinton Geoffrey.
\newblock Learning multiple layers of features from tiny images.
\newblock 2009.

\bibitem[Ankit et~al.(2019)Ankit, Hajj, Chalamalasetti, Agarwal, Marinella,
  Foltin, Strachan, Milojicic, mei Hwu, and Roy]{ankit2019panther}
Aayush Ankit, Izzat~El Hajj, Sai~Rahul Chalamalasetti, Sapan Agarwal, Matthew
  Marinella, Martin Foltin, John~Paul Strachan, Dejan Milojicic, Wen mei Hwu,
  and Kaushik Roy.
\newblock Panther: A programmable architecture for neural network training
  harnessing energy-efficient reram.
\newblock 2019.

\bibitem[Bulat(2019)]{flops}
Adrian Bulat.
\newblock pytorch estimate flops.
\newblock GitHub, 2019.

\bibitem[Denil et~al.(2013)Denil, Shakibi, Dinh, Ranzato, and
  de~Freitas]{redundancy}
Misha Denil, Babak Shakibi, Laurent Dinh, Marc’Aurelio Ranzato, and Nando
  de~Freitas.
\newblock Predicting parameters in deep learning.
\newblock In \emph{Advances in Neural Information Processing Systems 27}, pages
  2148--2156. 2013.

\bibitem[Dong et~al.(2017)Dong, Huang, Yang, and Yan]{dong_cvpr17}
Xuanyi Dong, Junshi Huang, Yi~Yang, and Shuicheng Yan.
\newblock More is less: A more complicated network with less inference
  complexity.
\newblock In \emph{Conference on Computer Vision and Pattern Recognition}.
  2017.

\bibitem[Garg et~al.(2020)Garg, Panda, and Roy]{Garg_2020}
Isha Garg, Priyadarshini Panda, and Kaushik Roy.
\newblock A low effort approach to structured cnn design using pca.
\newblock \emph{IEEE Access}, 8:\penalty0 1347–1360, 2020.
\newblock ISSN 2169-3536.
\newblock \doi{10.1109/access.2019.2961960}.

\bibitem[Guo et~al.(2016)Guo, Yao, and Chen]{guo_nips16}
Yiwen Guo, Anbang Yao, and Yurong Chen.
\newblock Dynamic network surgery for efficient dnns.
\newblock In \emph{Advances in Neural Information Processing Systems 29}. 2016.

\bibitem[Han et~al.(2015)Han, Pool, Tran, and Dally]{connections_nips15}
Song Han, Jeff Pool, John Tran, and William Dally.
\newblock Learning both weights and connections for efficient neural network.
\newblock In \emph{Advances in Neural Information Processing Systems 28}, pages
  1135--1143. 2015.

\bibitem[Han et~al.(2016)Han, Mao, and Dally]{compressing_iclr16}
Song Han, Huizi Mao, and William~J. Dally.
\newblock Deep compression: Compressing deep neural network with pruning,
  trained quantization and huffman coding.
\newblock In \emph{International Conference on Learning Representations}. 2016.

\bibitem[Han et~al.(2016a)Han, Liu, Mao, Pu, Pedram, Horowitz, and J]{Eie}
Song Han, Xingyu Liu, Huizi Mao, Jing Pu, Ardavan Pedram, Mark~A Horowitz, and
  William J.
\newblock Eie: efficient inference engine on compressed deep neural network.
\newblock In \emph{Computer Architecture (ISCA), 2016 ACM/IEEE 43rd Annual
  International Symposium}. 2016a.

\bibitem[He et~al.(2016)He, Zhang, Ren, and Sun]{res:2016}
Kaiming He, Xiangyu Zhang, Shaoqing Ren, and Jian Sun.
\newblock Deep residual learning for image recognition.
\newblock In \emph{Computer Vision and Pattern Recognition (CVPR)}. 2016.

\bibitem[He et~al.(2018)He, Kang, Dong, Fu, and Yang]{soft_2018}
Yang He, Guoliang Kang, Xuanyi Dong, Yanwei Fu, and Yi~Yang.
\newblock Soft filter pruning for accelerating deep convolutional neural
  networks.
\newblock In \emph{Proceedings of the Twenty-Seventh International Joint
  Conference on Artificial Intelligence (IJCAI-18)}, pages 2234--2240. 2018.

\bibitem[He et~al.(2019)He, Liu, Wang, Hu, and Yang]{He_2019_CVPR}
Yang He, Ping Liu, Ziwei Wang, Zhilan Hu, and Yi~Yang.
\newblock Filter pruning via geometric median for deep convolutional neural
  networks acceleration.
\newblock In \emph{The IEEE Conference on Computer Vision and Pattern
  Recognition (CVPR)}, June 2019.

\bibitem[He et~al.(2017)He, Zhang, and Sun]{channel_iccv}
Yihui He, Xiangyu Zhang, and Jian Sun.
\newblock Channel pruning for accelerating very deep neural networks.
\newblock In \emph{International Conference on Computer Vision}, pages
  1389--1397. 2017.

\bibitem[Hu et~al.(2016)Hu, Peng, Tai, and Tang]{hu_network}
Hengyuan Hu, Rui Peng, Yu-Wing Tai, and Chi-Keung Tang.
\newblock Network trimming: A data-driven neuron pruning approach towards
  efficient deep architectures.
\newblock In \emph{arXiv preprint arXiv:1607.03250}. 2016.

\bibitem[Jia et~al.(2009)Jia, Wei, Richard, Li-Jia, Kai, and Li]{imagenet}
Deng Jia, Dong Wei, Socher Richard, Li~Li-Jia, Li~Kai, and Fei-Fei Li.
\newblock Imagenet: A large-scale hierarchical image database.
\newblock In \emph{CVPR}, 2009.

\bibitem[Krizhevsky et~al.(2012)Krizhevsky, Sutskever, and
  Hinton]{NIPS2012_4824}
Alex Krizhevsky, Ilya Sutskever, and Geoffrey~E Hinton.
\newblock Imagenet classification with deep convolutional neural networks.
\newblock In \emph{Advances in Neural Information Processing Systems 25}, pages
  1097--1105. 2012.

\bibitem[Lebedev and Lempitsky(2016)]{lebedev_cvpr16}
Vadim Lebedev and Victor Lempitsky.
\newblock Fast convnets using group-wise brain damage.
\newblock In \emph{Conference on Computer Vision and Pattern Recognition}.
  2016.

\bibitem[Lee et~al.(2018)Lee, Ajanthan, and Torr]{lee2018snip}
Namhoon Lee, Thalaiyasingam Ajanthan, and Philip H.~S. Torr.
\newblock Snip: Single-shot network pruning based on connection sensitivity.
\newblock 2018.

\bibitem[Li et~al.(2017)Li, Kadav, Durdanovic, Samet, and Graf]{pruning_iclr17}
Hao Li, Asim Kadav, Igor Durdanovic, Hanan Samet, and Hans~Peter Graf.
\newblock Pruning filters for efficient convnets.
\newblock In \emph{International Conference on Learning Representations}. 2017.

\bibitem[Liu et~al.(2017)Liu, Li, Shen, Huang, Yan, and Zhang]{iccv17_slimming}
Zhuang Liu, Jianguo Li, Zhiqiang Shen, Gao Huang, Shoumeng Yan, and Changshui
  Zhang.
\newblock Learning efficient convolutional networks through network slimming.
\newblock In \emph{International Conference on Computer Vision}, pages
  2736--2744. 2017.

\bibitem[Liu et~al.(2019)Liu, Sun, Zhou, Huanh, and Darrell]{rethinking_iclr19}
Zhuang Liu, Mingjie Sun, Tinghui Zhou, Gao Huanh, and Trevor Darrell.
\newblock Rethinking the value of network pruning.
\newblock In \emph{International Conference on Learning Representations}. 2019.

\bibitem[Louizos et~al.(2018)Louizos, Welling, and Kingma]{christos_iclr18}
Christos Louizos, Max Welling, and Diederik~P Kingma.
\newblock Learning sparse neural networks through $l_0$ regularization.
\newblock In \emph{International Conference on Learning Representations}. 2018.

\bibitem[Luo et~al.(2017)Luo, Wu, and Lin]{thinnet}
Jian-Hao Luo, Jianxin Wu, and Weiyao Lin.
\newblock Thinet: A filter level pruning method for deep neural network
  compression.
\newblock In \emph{International Conference on Computer Vision}, pages
  5058--5066. 2017.

\bibitem[Molchanov et~al.(2017)Molchanov, Ashukha, and Vetrov]{dropout_icml17}
Dmitry Molchanov, Arsenii Ashukha, and Dmitry Vetrov.
\newblock Variational dropout sparsifies deep neural networks.
\newblock In \emph{International Conference on Machine Learning}. 2017.

\bibitem[{Molchanov} et~al.(2019){Molchanov}, {Mallya}, {Tyree}, {Frosio}, and
  {Kautz}]{cvpr_2019}
P.~{Molchanov}, A.~{Mallya}, S.~{Tyree}, I.~{Frosio}, and J.~{Kautz}.
\newblock Importance estimation for neural network pruning.
\newblock In \emph{2019 IEEE/CVF Conference on Computer Vision and Pattern
  Recognition (CVPR)}, pages 11256--11264, 2019.

\bibitem[Montavon et~al.(2017)Montavon, Lapuschkin, Binder, Samek, and
  Müller]{Montavon_2017}
Grégoire Montavon, Sebastian Lapuschkin, Alexander Binder, Wojciech Samek, and
  Klaus-Robert Müller.
\newblock Explaining nonlinear classification decisions with deep taylor
  decomposition.
\newblock \emph{Pattern Recognition}, 65:\penalty0 211–222, May 2017.
\newblock ISSN 0031-3203.
\newblock \doi{10.1016/j.patcog.2016.11.008}.

\bibitem[Paszke et~al.(2017)Paszke, Gross, Massa, Lerer, Bradbury, Chanan,
  Killeen, Lin, Gimelshein, Antiga, Desmaison, Köpf, Yang, DeVito, Raison,
  Tejani, Chilamkurthy, Steiner, Fang, Bai, and Chintala]{pytorch}
Adam Paszke, Sam Gross, Francisco Massa, Adam Lerer, James Bradbury, Gregory
  Chanan, Trevor Killeen, Zeming Lin, Natalia Gimelshein, Luca Antiga, Alban
  Desmaison, Andreas Köpf, Edward Yang, Zach DeVito, Martin Raison, Alykhan
  Tejani, Sasank Chilamkurthy, Benoit Steiner, Lu~Fang, Junjie Bai, and Soumith
  Chintala.
\newblock Pytorch: Tensors and dynamic neural networks in python with strong
  gpu acceleration.
\newblock 2017.

\bibitem[Prakash et~al.(2018)Prakash, Storer, Florencio, and
  Zhang]{prakash2018repr}
Aaditya Prakash, James Storer, Dinei Florencio, and Cha Zhang.
\newblock Repr: Improved training of convolutional filters.
\newblock In \emph{arXiv preprint arXiv:1811.07275}. 2018.

\bibitem[Roy et~al.(2020)Roy, Panda, Srinivasan, and Raghunathan]{roy2020}
Sourjya Roy, Priyadarshini Panda, Gopalakrishnan Srinivasan, and Anand
  Raghunathan.
\newblock Pruning filters while training for efficiently optimizing deep
  learning networks.
\newblock In \emph{arXiv preprint arXiv:2003.02800}. 2020.

\bibitem[Sebastian et~al.(2015)Sebastian, Alexander, Gregorie, Frederick,
  Klaus-Robert, and Wojciech]{lrp:2015}
Bach Sebastian, Binder Alexander, Montavon Gregorie, Klauschen Frederick,
  Muller Klaus-Robert, and Samek Wojciech.
\newblock On pixel-wise explanations for non-linear classifier decisions by
  layer-wise relevance propagation.
\newblock In \emph{Plos One}. 2015.

\bibitem[Wojciech et~al.(2016)Wojciech, Gregorie, Alexander, Sebastian, and
  Klaus-Robert]{lrp:2016}
Samek Wojciech, Montavon Gregorie, Binder Alexander, Lapuschkin Sebastian, and
  Muller Klaus-Robert.
\newblock Interpreting the predictions of complex ml models by layer-wise
  relevance propagation.
\newblock In \emph{arXiv preprint arXiv:1611.08191v1}. 2016.

\bibitem[Ye et~al.(2018)Ye, Lu, Lin, and Wang]{Ye_iclr18}
Jianbo Ye, Xin Lu, Zhe Lin, and James~Z Wang.
\newblock Rethinking the smaller-norm-less-informative assumption in channel
  pruning of convolution layers.
\newblock In \emph{International Conference on Learning Representations}. 2018.

\bibitem[You et~al.(2019)You, Yan, Ye, Ma, and Wang]{NIPS2019_gd}
Zhonghui You, Kun Yan, Jinmian Ye, Meng Ma, and Ping Wang.
\newblock Gate decorator: Global filter pruning method for accelerating deep
  convolutional neural networks.
\newblock In \emph{Advances in Neural Information Processing Systems 32}, pages
  2133--2144. Curran Associates, Inc., 2019.

\bibitem[Yu et~al.(2018)Yu, Li, Chen, Lai, Morariu, Han, Gao, Lin, and
  Davis]{nisp}
Ruichi Yu, Ang Li, Chun-Fu Chen, Jui-Hsin Lai, Vlad~I. Morariu, Xintong Han,
  Mingfei Gao, Vhing-Yung Lin, and Larry~S. Davis.
\newblock Nisp: Pruning networks using neuron importance score propagation.
\newblock In \emph{IEEE/CVF Conference on Computer Vision and Pattern
  Recognition}. 2018.

\bibitem[Yue et~al.(2019)Yue, Weibin, and Lin]{yue2019really}
Li~Yue, Zhao Weibin, and Shang Lin.
\newblock Really should we pruning after model be totally trained? pruning
  based on a small amount of training.
\newblock In \emph{arXiv preprint arXiv:1901.08455}. 2019.

\bibitem[Zagoruyko(2015)]{vgg16}
Sergey Zagoruyko.
\newblock 92.45\% on cifar-10 in torch.
\newblock In \emph{http://torch.ch/blog/2015/07/30/cifar.html}. 2015.

\bibitem[Zhuang et~al.(2018)Zhuang, Tan, Zhuang, Liu, Guo, Wu, Huang, and
  Zhu]{dcp_nips18}
Zhuangwei Zhuang, Mingkui Tan, Bohan Zhuang, Jing Liu, Yong Guo, Qingyao Wu,
  Junzhou Huang, and Jinhui Zhu.
\newblock Discrimination-aware channel pruning for deep neural networks.
\newblock In \emph{Advances in Neural Information Processing Systems 31}, pages
  875--886. 2018.

\end{thebibliography}

\pagebreak
\section*{Supplementary Material}
\subsection*{$\alpha \beta$-Decomposition Rule}
Layer-wise relevance propagation (LRP) \cite{lrp:2015} builds a local redistribution rule for each node of the network, and applies these rules in a backward pass in order to produce the pixel-wise decomposition.
It allocates a relevance score to each node in the network using the activations and weights of the network for an input image.
In particular, LRP associates a vector of scores to each node indication how relevant they are to the prediction.
Various rules has been proposed in the literature to redistribute the relevance scores assigned to the final layer onto the input nodes \cite{Montavon_2017, lrp:2016}. Note, sensitivity analysis and decomposition methods such as LRP are qualitatively different as explained in \cite{lrp:2016}. 

The $\alpha \beta$-decomposition rule redistributes the relevance scores from layer $l+1$ to layer $l$ as shown in equation.~\ref{eq:1}, where $()_+$ indicates the positive weight components and $()_-$ indicates the negative weight components. The relevance at each layer is conserved by enforcing $\alpha-\beta=1$. If we set $\alpha=1$ and $\beta=0$ then it is same as the $z_+$ redistribution rule proposed in \cite{Montavon_2017} which only considers the positive weights. It was shown in \cite{Montavon_2017} that the $z_+$ redistribution rule results from `Deep Taylor Decomposition' of the neural network function when the non-linearity neurons used in the network are ReLUs. 

\begin{equation}
\label{eq:1}
    R_i^l = \sum_{j}(\alpha \frac{(a_i w_{ij})^{+}}{\sum_i {(a_i w_{ij})^{+}}} -\beta\frac{(a_i w_{ij})^{-}}{\sum_i {(a_i w_{ij})^{-}}})R_j^{l+1}
\end{equation}

\section*{Implementation Details}
We have implemented the proposed technique on PyTorch \cite{pytorch}. We have used the Stochastic Gradient Descent (SGD) algorithm with momentum and weight decay to train the networks. For the CIFAR-10 and CIFAR-100 datasets, the momentum and weight decay are set as 0.9 and 0.0005, respectively. 
\begin{table}[ht]
\caption{Test accuracy, number of parameters and number of Floating Point Operations (FLOPs) required for inference of a baseline (Unpruned) model.}
\label{tab:baseline}
\vskip 0.15in
\begin{center}
\begin{small}
\begin{tabular}{|l|c|c|c|c|}
\hline
Dataset& Model & Parameters (in Millions) & FLOPS (x$10^9$) & Accuracy (\%)\\
\hline
  & VGG-16& 14.99&0.32 &93.95\\
  CIFAR-10& ResNet-56&0.86 & 0.13& 93.61\\
  &ResNet-110& 1.73&0.26 & 93.86\\
  &ResNet-164& 1.70 &0.26 & 94.80\\
 \hline
 & ResNet-56&0.86 & 0.13& 71.48\\
  CIFAR-100&ResNet-110&1.74 & 0.26&73.28\\
  &ResNet-164& 1.73&0.26 &75.73\\
 \hline
ImageNet& ResNet-34 & 21.79&14.29 & 75.12\\
  \hline
 \end{tabular}
\end{small}
\end{center}
\vspace{-0.5cm}
\end{table}
For ImageNet we have used 0.9 as momentum and 0.0001 as weight decay. 
The models for CIFAR-10 and CIFAR100 datasets are trained for $N=200$ epochs with a batch size of 256. 
The initial learning rate is set at 0.1 and is divided by 10 at epochs 100 and 150. 
The models for ImageNet are trained for $N=65$ epochs with a batch size of 64. 
The initial learning rate is set at 0.01 and is divided by 10 after every 30 epochs. The value of $N1$ (see algorithm 2) is fixed at 150 for CIFAR-10, CIFAR-100 and 40 for ImageNet. The hyper-parameter $n$ typically varies around $(10-20)$ for CIFAR datasets and is set as $9$ for ImageNet. 
The input data is normalized using channel means and standard deviations.
The following transforms are applied to the training data: \textit{transforms.RandomCrop} and \textit{transforms.RandomHorizontalFlip} \cite{pytorch}.

We have a used modified version of VGG-16 \cite{vgg16} with batch-normalization which has fewer parameters in the linear layers. The total number of parameters in the unpruned VGG-16 network is 15 million. The baseline test accuracy, number of parameters and Floating Point Operations (FLOPs) required for inference for all the datasets and architectures is shown in Table.~\ref{tab:baseline}.
We have adopted the PyTorch utility that estimates the number of FLOPs for a given network presented in \cite{flops}.

\end{document}